\documentclass{article}
\usepackage{spconf,amsmath,graphicx}

\usepackage[table]{xcolor}
\usepackage{adjustbox}
\usepackage{graphicx}
\usepackage{float}
\usepackage{amsfonts}
\usepackage{booktabs,caption}
\usepackage[flushleft]{threeparttable}
\usepackage{float}
\usepackage{placeins}

\usepackage[T1]{fontenc}
\usepackage{newtxtext,newtxmath}

\title{4D Neural Voxel Splatting: Dynamic Scene Rendering with Voxelized Guassian Splatting}
\name{Chun-Tin Wu$^1$, Jun-Cheng Chen$^2$}
\address{$^1$National Taiwan University, $^2$Academia Sinica
\thanks{This research is supported by the National Science and Technology Council (NSTC), Taiwan under Grants 114-2221-E-001-004, 114-2221-E-001-016, 114-2634-F-001-001-MBK, and 114-2634-F-002-004 by Academia Sinica under Grants AS-IAIA-114-M10 and AS-GCS-115-M06. We thank to National Center for High-performance
Computing (NCHC) of National Applied Research Laboratories (NARLabs) in Taiwan for providing computational
and storage resources.}}

\begin{document}
\ninept
\maketitle

\begin{abstract}
Although 3D Gaussian Splatting (3D-GS) achieves efficient rendering for novel view synthesis, extending it to dynamic scenes still results in substantial memory overhead from replicating Gaussians across frames. To address this challenge,  we propose 4D Neural Voxel Splatting (4D-NVS), which combines voxel-based representations with neural Gaussian splatting for efficient dynamic scene modeling. Instead of generating separate Gaussian sets per timestamp, our method employs a compact set of neural voxels with learned deformation fields to model temporal dynamics. The design greatly reduces memory consumption and accelerates training while preserving high image quality. We further introduce a novel view refinement stage that selectively improves challenging viewpoints through targeted optimization, maintaining global efficiency while enhancing rendering quality for difficult viewing angles. Experiments demonstrate that our method outperforms state-of-the-art approaches with significant memory reduction and faster training, enabling real-time rendering with superior visual fidelity.
\end{abstract}

\vspace{-10pt}
\section{Introduction}
\label{sec:intro}

Dynamic 3D scene reconstruction and novel view synthesis are recent research hotspots 
in computer vision with a widespread applications in virtual reality (VR), augmented reality (AR), and digital entertainment. Capturing and rendering temporal phenomena—such as human motion or deformable objects—is essential for immersive experiences.  

While Neural Radiance Fields (NeRF) \cite{nerf} achieve photorealistic view synthesis in static scenes, extending them to dynamics faces key bottlenecks: long training times, slow inference, and poor scalability with temporal complexity. Acceleration methods like DirectVoxGO \cite{directvoxgo} and Instant-NGP \cite{instant-ngp} improve efficiency but remain limited by volumetric rendering. 3D Gaussian Splatting (3D-GS) \cite{3dgs} shifts toward explicit point-based rendering, enabling real-time performance, yet naive extensions to dynamics suffer from memory growth proportional to sequence length. Canonical-space methods like 4D-GS \cite{4dgs} reduce this but rely on heavy deformation networks and costly backward mapping. Recent advances attempt to balance efficiency and expressiveness: Ex4DGS \cite{ex4dgs} improves temporal consistency via explicit motion modeling, FreeTimeGS \cite{freetimegs} decouples space and time for flexible representation, and LongVolCap \cite{longvolcap} compresses long sequences effectively. 
%
%
However, the practical deployment of these methods remains constrained by memory and training overhead, especially in the scenarios of embodied AI which has a critical demand for the resource consumption. For instance, 
Mobile robots on edge devices (e.g., NVIDIA Jetson with 4--8GB memory) must rapidly reconstruct and track dynamic scenes for navigation and interaction. Field robots likewise require scene adaptation within minutes, not hours, to operate effectively. Current methods often exceed available resources on edge or mid-range GPUs, forcing compromises in scene complexity or hardware cost. Thus, the core challenge is \textit{how to capture rich temporal dynamics while keeping memory and computational efficiency for real-time, resource-constrained applications.}\\
\indent To address these limitations, we introduce \emph{4D Neural Voxel Splatting (4D-NVS)}, which combines voxel-based efficiency with Gaussian splatting quality through learned temporal deformation. Our key insight is decoupling spatial structure from temporal dynamics: we generate neural Gaussians on-demand from persistent voxel anchors and apply unified deformations. This achieves $\mathcal{O}(fV+F)$ memory complexity compared to traditional $\mathcal{O}(N \cdot T)$ scaling, where $V$ is the number of voxels, $f$ is the feature dimension per voxel, $F$ represents the deformation network parameters, $N$ is the number of Gaussians, and $T$ is the number of timestamps.s While our method builds upon established techniques (voxel-based generation from Scaffold-GS \cite{scaffoldgs} and HexPlane decomposition \cite{hexplane}), we introduce several key innovations that distinguish our approach:
\begin{itemize}
    \item \textbf{Unified 4D Voxel Architecture:} We extend 3D voxel grids to 4D by treating time as an additional dimension in the voxel feature space, enabling temporal-aware Gaussian generation that adapts both spatially and temporally. Unlike Scaffold-GS which generates static Gaussians, our voxels produce time-varying Gaussians through learned temporal features.
    
    \item \textbf{Selective Deformation Strategy:} Through extensive experimentation, we identified that deforming all Gaussian properties leads to training instability. We introduce a selective approach that only deforms geometric properties (position, scale, rotation) while keeping appearance properties (color, opacity) fixed, greatly improving convergence and quality.
    
    \item \textbf{View-Adaptive Refinement:} We propose a novel refinement mechanism that identifies and selectively improves underperforming viewpoints through adaptive densification, addressing temporal inconsistencies without global overhead.

    \item \textbf{Memory-Efficient Design:} Our framework achieves $\mathcal{O}(fV+F)$ memory complexity instead of $\mathcal{O}(N\cdot T)$. This makes dynamic scene rendering feasible on consumer GPUs.

\end{itemize}

To sum up, the proposed approach successfully bridges computational efficiency and visual quality, enabling new possibilities for real-time dynamic visualization and interactive scene manipulation.

\begin{figure*}[t]
    \centering
    \includegraphics[width=1\textwidth]{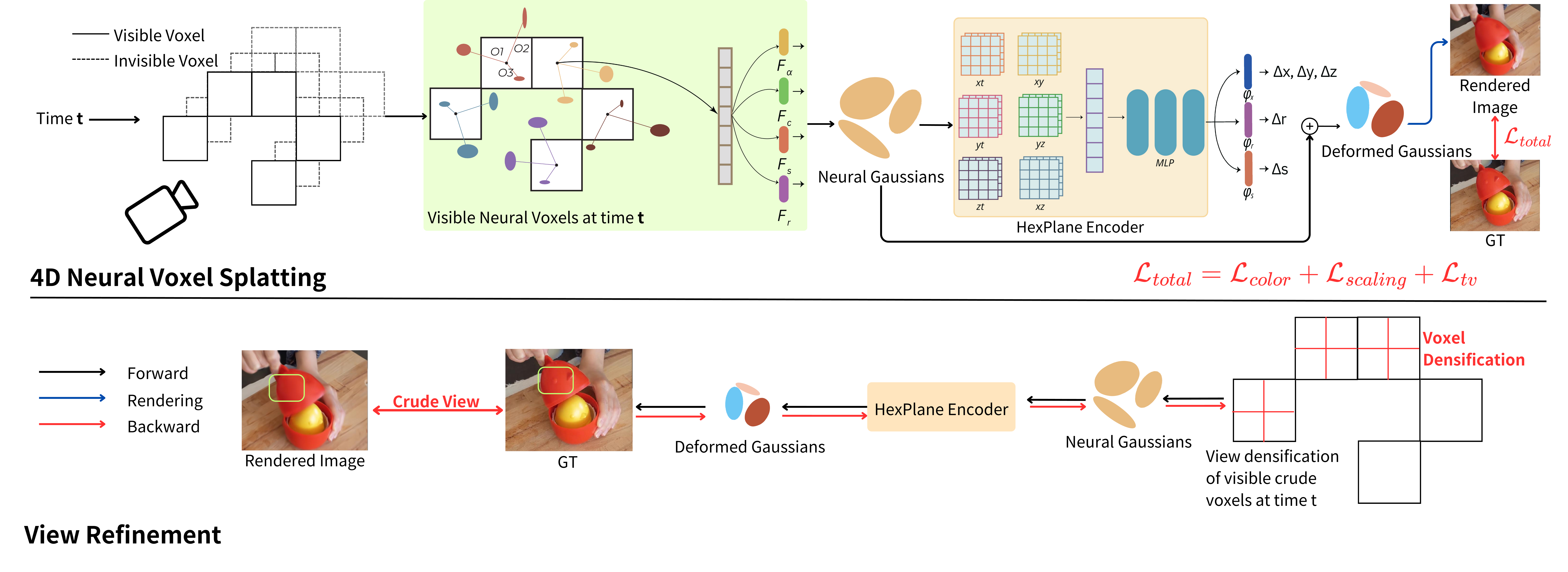}
    \vspace{-20pt}
    \caption{ 
    Pipeline overview: (1) Initialize with voxel-based Gaussian splatting, (2) Generate neural Gaussians with temporal information, (3) Apply HexPlane temporal corrections, (4) Optimize with color loss, total variation loss, and scaling regularization, and (5) View refinement stage for underperforming viewpoints through adaptive densification.}
    \label{fig:Overview}
    \vspace{-10pt}
\end{figure*}

\section{Preliminaries}
\label{sec:prelim}

\subsection{3D Gaussian Splatting}
3D Gaussian Splatting (3D-GS) \cite{3dgs} represents scenes using anisotropic 3D Gaussians, each defined by a mean position $\mu \in \mathbb{R}^3$ and covariance matrix $\Sigma$:
\begin{equation} 
G(x) = e^{-\frac{1}{2}(x-\mu)^\top \Sigma^{-1} (x-\mu)},
\end{equation} 
\noindent where the covariance matrix is constructed as $\Sigma = R S S^T R^T$ using scaling matrix $S$ and rotation matrix $R$. Each Gaussian has associated color $c$ and opacity $\alpha$ (modeled with spherical harmonics) .

Unlike volumetric rendering, 3D-GS projects Gaussians onto the image plane as 2D Gaussians and applies $\alpha$-blending:
\begin{equation} 
C(x') = \sum_{i=1}^{N} c_i \alpha_i \prod_{j=1}^{i-1} (1 - \alpha_j),
\end{equation} 
where $x'$ is the pixel position, $\alpha$ depends on the pixel position $x'$ and $N$ is the number of overlapping Gaussians. This tile-based rasterization enables real-time rendering with differentiable optimization.

\subsection{Scaffold-GS: Grid-Based Gaussian Generation}
Scaffold-GS \cite{scaffoldgs} generates Gaussians from structured anchor points rather than optimizing individual primitives. Starting from a sparse point cloud $P \in \mathbb{R}^{M \times 3}$, it creates voxel centers:
\begin{equation}
\mathbf{V} = \left\{ \left\lfloor \frac{\mathbf{P}}{\epsilon} \right\rceil \right\} \cdot \epsilon,
\end{equation}
where $\epsilon$ is the voxel size. Each voxel center $v \in V$ serves as an anchor with local features $f_v$, scaling factor $l_v \in \mathbb{R}^3$, and $k$ learnable offsets $O_v \in \mathbb{R}^{k \times 3}$.

Neural Gaussians are generated on-demand within the viewing frustum. For each visible anchor, $k$ Gaussians are spawned with positions $\mu_i = x_v + O_i \cdot l_v$, while attributes (opacity, color, scale, rotation) are decoded from anchor features using MLPs conditioned on viewing direction and distance. This approach significantly reduces memory requirements compared to storing explicit Gaussians for the entire scene.
\section{Method}
\label{sec:method}
Our 4D Neural Voxel Splatting (4D-NVS) framework addresses the challenge of efficiently rendering dynamic scenes by extending traditional 3D voxel representations into the temporal domain. 
As shown in Fig.~\ref{fig:Overview}, we achieve this through a three-stage training approach: (1) \textbf{coarse initialization} builds a foundational geometric representation from mixed timestamps without modeling temporal deformation, (2) \textbf{fine temporal training} enables the HexPlane deformation module to learn full spatiotemporal dynamics while identifying poorly reconstructed viewpoints, and (3) \textbf{view refinement} selectively re-trains only those challenging viewpoints with stronger densification to improve quality without degrading already well-reconstructed areas. The details are described in the following sections, respectively.


%

\subsection{Initialization and Voxel Setup}
We initialize neural voxels using the sparse point cloud from Structure-from-Motion (SfM) \cite{colmap}, distributing voxel centers to ensure comprehensive spatial coverage. The scene is first trained using Scaffold-GS \cite{scaffoldgs} for 3K iterations with mixed timestamp samples to establish a foundational geometric representation.

\subsection{Neural Gaussian Generation}

\noindent\textbf{On-Demand Generation.} Unlike existing methods that pre-compute and store Gaussians, our approach generates neural Gaussians dynamically based on viewing conditions and temporal context. This represents a fundamental shift from static storage to dynamic computation, enabling both memory efficiency and view-dependent optimization.
At each timestamp $t$, we perform visibility culling to identify voxels within the camera frustum, significantly reducing computational overhead from $\mathcal{O}(N)$ to $\mathcal{O}(V_{visible})$ where $V_{visible} \ll N$. For each visible voxel $v$, we generate $k$ neural Gaussians with positions:
\begin{equation}
\mu_i = x_v + O_i \cdot l_v, \quad i = 0, \ldots, k-1,
\end{equation}
where $x_v \in \mathbb{R}^3$ is the voxel center position, $O_i \in \mathbb{R}^3$ are learnable offset vectors for each of the $k$ Gaussians, and $l_v \in \mathbb{R}^3$ is the voxel's spatial extent (scale factor).

Gaussian attributes are decoded using dedicated MLPs that take anchor features $f_v$, viewing distance $\delta_{vc}$, and directional vector $\vec{d}_{vc}$ as input:
\begin{align}
\{ \alpha_0, \ldots, \alpha_{k-1} \} &= F_\alpha(f_v, \delta_{vc}, \vec{d}_{vc}), \\
\{ c_0, \ldots, c_{k-1} \} &= F_c(f_v, \delta_{vc}, \vec{d}_{vc}), \\
\{ s_0, \ldots, s_{k-1} \} &= F_s(f_v, \delta_{vc}, \vec{d}_{vc}), \\
\{ r_0, \ldots, r_{k-1} \} &= F_r(f_v, \delta_{vc}, \vec{d}_{vc}),
\end{align}
where $F_\alpha$, $F_c$, $F_s$, and $F_r$ are MLPs for opacity, color, scale, and rotation respectively.
\subsection{Dynamic Gaussians Deformation}
\noindent\textbf{Unified 4D Representation.} Our method introduces a novel temporal modeling approach that treats space and time as a unified 4D manifold while maintaining computational efficiency. Unlike previous methods that either use separate deformation networks for each Gaussian or employ expensive per-frame optimization, we leverage a shared HexPlane decomposition that captures temporal correlations across the entire scene.

To capture dynamic motihon, we employ HexPlane \cite{kplanes} decomposition, encoding 4D space into six planes with multi-resolution temporal features $f_h$. Specifically, HexPlane factorizes the 4D space-time volume into six 2D planes: three spatial planes (XY, XZ, YZ) and three space-time planes (XT, YT, ZT). For a 4D point $(x, y, z, t)$, we extract features via bilinear interpolation from each plane and aggregate them:
\begin{equation}
f_h = \sum_{p \in \{\text{XY}, \text{XZ}, \text{YZ}, \text{XT}, \text{YT}, \text{ZT}\}} \text{Interp}_p(x, y, z, t).
\end{equation}
\noindent This decomposition provides several advantages over direct 4D parameterization: (1) significantly reduced memory footprint (2) natural handling of temporal correlations, and (3) efficient gradient flow during optimization. A compact MLP $\phi_d$ integrates these features:
\begin{equation}
f_d = \phi_d(f_h),
\end{equation}
where $\phi_d$ is a shallow network for computational efficiency.
Separate deformation decoders predict position, rotation, and scale changes:
\begin{align}
\Delta \mu = \varphi_x(f_d),\quad \Delta r = \varphi_r(f_d),\quad \Delta s = \varphi_s(f_d) 
\end{align}
where $\varphi_x$, $\varphi_r$, and $\varphi_s$ are single-layer MLPs that output 3D position offsets, quaternion rotations, and 3D scale factors, respectively.

The final deformed Gaussian values are $(\mu', r', s') = (\mu + \Delta \mu, r + \Delta r, s + \Delta s)$, while color and opacity remain unchanged to prevent error propagation, a design choice that emerged from our analysis of training stability in dynamic scenarios.

\noindent\textbf{Selective Deformation Strategy.} Through extensive experimentation, we discovered that deforming all five Gaussian properties (position, opacity, color, scale, rotation) leads to training instability and error accumulation. Specifically, we found that appearance properties (color and opacity) are highly sensitive to deformation errors, causing cascading failures during backpropagation. By keeping these properties fixed and only deforming geometric attributes, we maintain stable gradients while still capturing complex motion patterns.

\subsection{Optimization}
Our loss function combines multiple terms for robust optimization:
\begin{equation}
\mathcal{L} = \mathcal{L}_{color} + \lambda_{tv} \mathcal{L}_{tv} + \lambda_{vol} \mathcal{L}_{vol},
\end{equation}
where $\mathcal{L}_{color} = \mathcal{L}_1 + \lambda_{SSIM} \mathcal{L}_{SSIM}$ ensures pixel accuracy and structural coherence, $\mathcal{L}_{tv}$ enforces spatial smoothness in the HexPlane, and $\mathcal{L}_{vol} = \sum_{i=1}^{N_{ng}} \mathbf{s_x}_i\cdot\mathbf{s_y}_i\cdot\mathbf{s_z}_i$ prevents oversized Gaussians.

For density control, we grow anchors in high-gradient regions and prune those producing consistently low-opacity Gaussians.

\subsection{View Refinement}
\noindent\textbf{Motivation.} During training, we observed that certain viewpoints consistently under-perform due to large deformations, complex occlusions, or rapid temporal changes. Rather than applying uniform densification across all views—which wastes computational resources on already well-reconstructed areas—we developed an adaptive refinement strategy that identifies and specifically improves problematic viewpoints while maintaining global efficiency.

\noindent\textbf{Crude View Selection.}
We implement two complementary strategies for identifying crude viewpoints that require refinement:

\noindent \textbf{1. PSNR-based Detection.} We track the PSNR of each rendered view and compare it against an exponentially weighted moving average (EMA) to identify statistical outliers:
\begin{equation}
\text{PSNR}_i < (1+\gamma) \cdot \text{EMA}_{\text{PSNR}},
\end{equation}
where $\gamma$ starts at 0.05 and gradually decays to 0.02 during training to become more selective as optimization progresses. The EMA is updated with momentum 0.4 to balance responsiveness and stability:
\begin{equation}
\text{EMA}_{\text{PSNR}} = 0.4 \cdot \text{PSNR}_{\text{current}} + 0.6 \cdot \text{EMA}_{\text{PSNR}}.
\end{equation}

\noindent \textbf{2. Gradient-based Detection.} For scenarios where gradient information reveals optimization difficulties not captured by quality metrics, we track the gradient magnitude of viewspace points and identify views with abnormally high gradients indicating convergence issues:
\begin{equation}
\|\nabla \mathcal{L}_{\text{view}}\| > (1+\gamma) \cdot \text{EMA}_{\text{grad}}.
\end{equation}
This dual-criteria approach ensures robust detection across different failure modes—quality-based detection captures rendering artifacts while gradient-based detection identifies optimization instabilities.

Identified viewpoints are added to a refinement stack with associated metadata including failure type (quality vs. gradient), severity score, and temporal consistency flags. Views appearing consecutively in multiple frames receive higher priority for refinement.

\noindent\textbf{Adaptive Quality Enhancement.} 
For crude views, we apply specialized training in a dedicated third stage (14k iterations) that focuses computational resources exclusively on problematic areas:

\noindent \textbf{1. Focused Training.} Only crude viewpoints from the adaptive camera list are used for training, with sampling probability weighted by severity scores. This concentrates gradient updates on areas that need improvement most, avoiding dilution of learning signals from well-performing regions.

\noindent \textbf{2. Adjusted Thresholds.} We use more aggressive densification parameters specifically tuned for challenging views:
These lower thresholds encourage more frequent Gaussian splitting and pruning in problematic regions.

\noindent \textbf{3. Enhanced Gaussian Generation.} The reduced thresholds facilitate creation of additional Gaussians in regions with complex motion patterns, fine-grained occlusions, or temporal discontinuities. New Gaussians inherit temporal features from nearby anchors and undergo immediate deformation field training to capture local dynamics.

This targeted approach enhances overall quality and temporal consistency without computational overhead on well-performing areas, achieving 0.5-1.2 dB PSNR improvements on challenging viewpoints while maintaining global rendering efficiency.
\section{Experiments}
\label{sec:experiments}
\begin{table}[t]
\centering
\resizebox{1\linewidth}{!}{%
\begin{threeparttable}
\begin{tabular}{@{}lcccccc@{}}
\toprule
Model & PSNR(dB) \(\uparrow\) & MS-SSIM\(\uparrow\) & Times\(\downarrow\) & FPS\(\uparrow\) & Memory (MB)
\(\downarrow\) \\
\midrule
Nerfies \cite{nerfies} & 22.2 & 0.803 & $\sim$ hours & $<$ 1 & -\\
3D-GS* \cite{3dgs} & 19.7 & 0.68 & 40 mins & \cellcolor{red!50}55 & - \\
Scaffold-GS* \cite{scaffoldgs} & 20.7 & 0.688 & 35 mins & \cellcolor{red!50}55 & - \\
HyperNeRF \cite{hypernerf} & 22.4 & 0.814 & 32 hours & $<$ 1 & -\\
TiNeuVox-B \cite{tineuvox} & 24.3 & 0.836 & 18 mins** & 1 & 21,558** \\
\midrule
4D-GS \cite{4dgs} & 25.2 & 0.845 &  25 mins** & 34 & \cellcolor{yellow} 4,500**\\
GAGS \cite{gags} &24.26& 0.83 & 120 mins** & 11 & 6,875** \\
Ours w/o Refinement & \cellcolor{yellow}25.8 & \cellcolor{yellow}0.846 & \cellcolor{red!50}10 mins{\color{blue}†} & \cellcolor{yellow}45 & \cellcolor{red!50}3,050 \\
Ours & \cellcolor{red!50}28.5 & \cellcolor{red!50}0.872 & \cellcolor{yellow}13 mins{\color{blue}†} & 44 & \cellcolor{red!50}3,050 \\
\bottomrule
\end{tabular}
\begin{tablenotes}
        \footnotesize
        \centering
        \item * : The methods are  trained on randomly sampled timestamps from all frames, without temporal modeling.
        \item **: The training time and memory is re-evaluated by us on a single RTX 4090 GPU.
        \item †: Training time includes 3k initialization iterations (approximately 2 minutes).
\end{tablenotes}

\end{threeparttable}
}
\caption{Quantitative results on the HyperNeRF \cite{hypernerf} VRIG dataset, rendered at a resolution of 960$\times$540. The \colorbox{red!50}{best} and \colorbox{yellow}{second best} results are highlighted in red and yellow, respectively. Our method performed the best compared to other methods, while using the least memory and training time.}

\label{table:hypernerf}
\vspace{-10pt}
\end{table}

\begin{table}[t]
    \centering
    \resizebox{1\linewidth}{!}{%
    \scriptsize
    \begin{tabular}{@{}lccccc@{}}
    \toprule
    Model & PSNR(dB) \(\uparrow\) & D-SSIM\(\downarrow\) & Time\(\downarrow\) & FPS\(\uparrow\)  \\
    \midrule
    NeRFPlayer \cite{nerfplayer} & 30.69 & 0.034 &  6 hours & 0.045&  \\
    HyperReel \cite{hyperreel} & 31.10 & 0.036 &  9 hours & 2.0  &\\
    HexPlane \cite{hexplane} & 31.70 & \cellcolor{red!50}0.014 & 12 hours & 0.2 & \\
    KPlanes \cite{kplanes} & 31.63 & - &  1.8 hours & 0.3  & \\
    Im4D \cite{im4d} & \cellcolor{yellow}32.58 & - & 28 mins & ~5 & \\
    MSTH \cite{MSTH} & 32.37 & \cellcolor{yellow}0.015 & \cellcolor{red!50}20 mins & 2 & \\
    \midrule
    4D-GS{\color{blue} (CVPR'24)} \cite{4dgs} & 31.15 & 0.016 &  32 mins & 34 & \\
    4D-GS{\color{blue} (ICLR'24)} \cite{4dgs-iclr} & 31.91 & 0.015 & 105 mins & \cellcolor{red!50}114 & \\
    Ex4DGS\cite{ex4dgs}*& 32.11 & 0.015 & 36 mins & 25 & \\
    Ours** & {\color{blue}\cellcolor{red!50}33.12} & 0.021 &  \cellcolor{yellow}25 mins & \cellcolor{yellow}43 & \\
    \bottomrule
    \end{tabular}
    }
    \begin{tablenotes}
        \footnotesize
        \item * :The training time and FPS is evaluated by us on a single RTX-4090.
        \item ** :We introduced gradient-selection selection in view refinement stage\\ which improves the metrics.
        \vspace{-5pt}
    \end{tablenotes}
    \caption{Quantitative results on the Neu3D \cite{neu3d} dataset. Note: rendering resolution is set to 1,352$\times$1,014.}
    \label{table:neu3d}
    \vspace{-10pt}
\end{table}
\subsection{Experimental Setup}
\noindent\textbf{Setup.} We set \( k = 10 \) Gaussians per Neural Voxel and feature dimension \( f_v \in \mathbb{R}^{32} \). HexPlane grids use resolution $[64, 64, 64, 150]$ for DyNeRF and $[64, 64, 64, 25]$ for HyperNeRF with scale multipliers $[1, 2]$. Training consists of 3k initialization iterations, 14k main training, and 14k view refinement iterations.

\noindent\textbf{Implementation Details.}
We use Adam optimizer with learning rates: Gaussian offsets (0.01), color MLP (0.008), opacity MLP (0.002), rotation/scaling MLPs (0.004), HexPlane grids (0.0016), deformation MLPs (0.00016). All rates decay to 1/10 initial values by iteration 14,000.Loss weights: $\lambda_{vol} = 0.015$ for volume regularization and $\lambda_{tv} = 0.0002$ for total variation loss. Densification threshold: $\tau_{g} = 0.0002$ (coarse) to $0.0001$ (refinement); opacity threshold: $\tau_\alpha = 0.05$ to $0.03$.

\noindent\textbf{Dataset.} HyperNeRF \cite{hypernerf}: 1-2 cameras with straightforward motion. Neu3D \cite{neu3d}: 15-20 static cameras with complex motion. COLMAP initialization from first frame (Neu3D) or 200 random frames (HyperNeRF), yielding 5,000-20,000 initial points for voxel grid initialization. 
\subsection{Evaluation Results}
\noindent\textbf{Quantitative Results.}
To assess the quality of novel view synthesis, we conducted benchmarking against several state-of-the-art methods in the field. The results are summarized in Table \ref{table:hypernerf} and Table \ref{table:neu3d}. On HyperNeRF, our method achieves 28.5 PSNR while using the least memory (3,050 MiB) and fastest training time (13 minutes). On Neu3D, we maintain competitive quality (33.12 PSNR) with real-time rendering at 43 FPS—an order of magnitude faster than Im4D (5 FPS) and MSTH (2 FPS). This combination of high quality, minimal memory footprint, and real-time performance makes our approach ideal for practical deployment.

\begin{figure}[t]
    \centering
    \includegraphics[width=\linewidth]{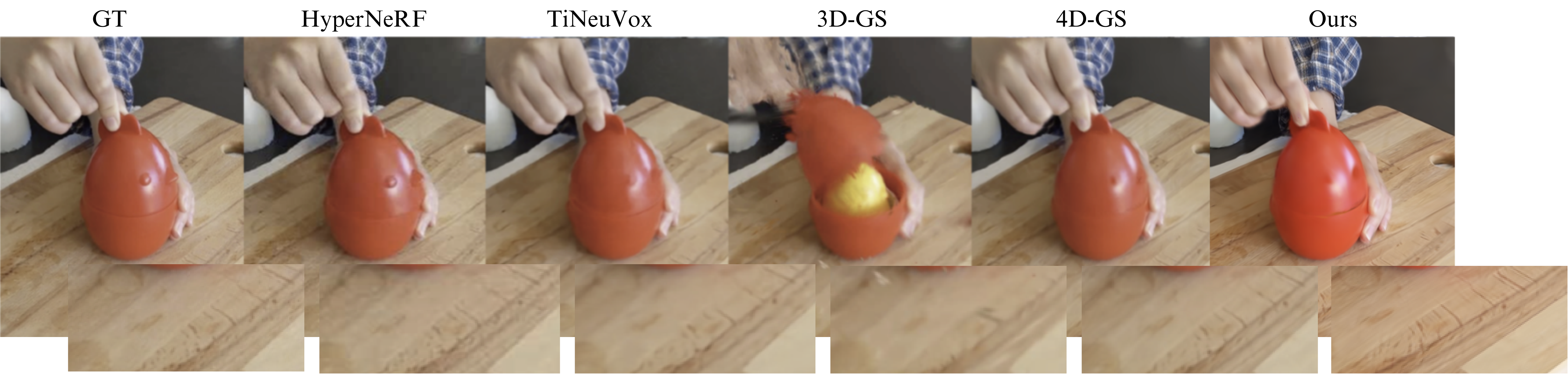}
    \caption{Visual comparisons of the proposed method on the HyperNeRF dataset with other methods. The proposed method achieves better rendering results. Best viewed at 4$\times$ zoom.}
    \label{fig:qualitative}
    \vspace{-10pt}
\end{figure}

\begin{figure}[t]
    \centering
    \includegraphics[width=1\linewidth]{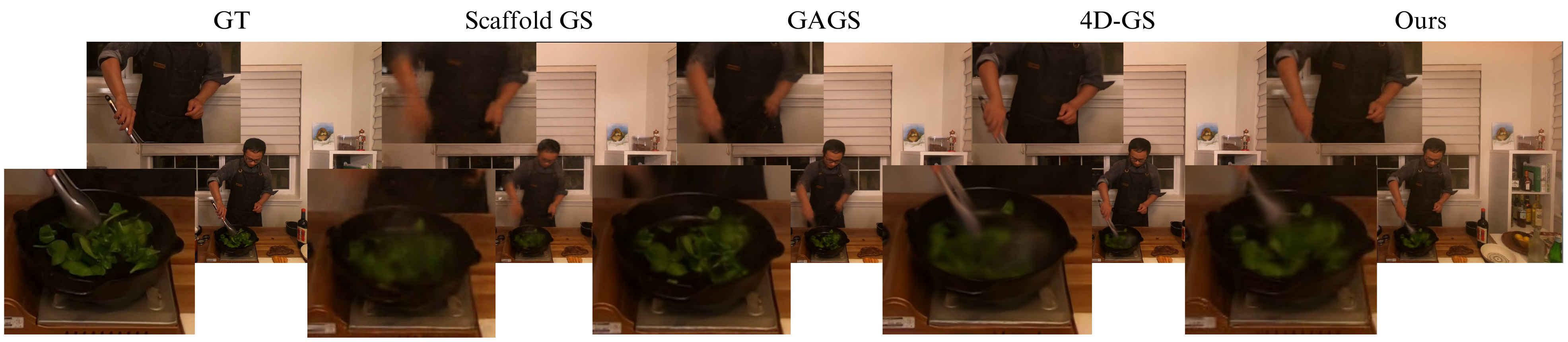}
    \caption{Visualization of the Neu3D dataset compared with other methods. From the visual illustration shown in the top and bottom left, the proposed method strikes a balance while the others either perform worse on the hand or the spinach in the pan. More rendering videos can be found in the supplementary materials. Best viewed at 4$\times$ zoom.}
    \label{fig:qualitative_neu3d}
    \vspace{-10pt}
\end{figure}

\begin{figure}[t]
    \centering
    \includegraphics[width=1\linewidth]{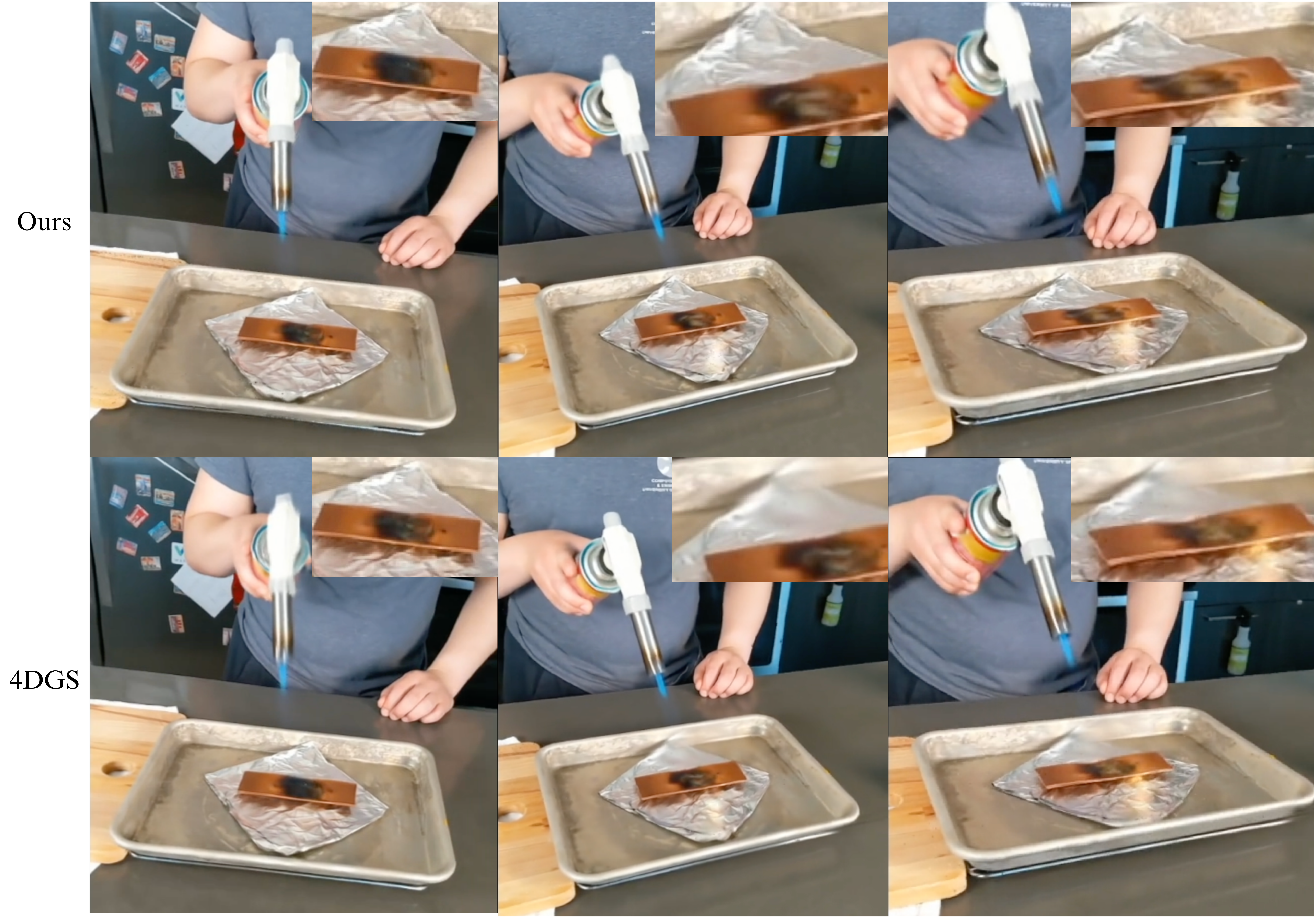}
    \vspace{-5pt}
    \caption{Continuous Frames on HyperNeRF Dataset compared with 4DGS.\textit{ \textbf{Top: Ours, Bottom: 4DGS.}} The proposed method deliver a better rendering results with more details than 4DGS, which can be seen in the top right. Best viewed at 4$\times$ zoom.}
    \label{fig:multi-frame}
    \vspace{-5pt}
\end{figure}

\noindent\textbf{Qualitative Results.}
The qualitative results of our study, compared with several other methods, are shown in Fig. \ref{fig:qualitative} and Fig. \ref{fig:qualitative_neu3d}. These results demonstrate the effectiveness of our approach. With reduced training time, rendering time, and memory consumption, our method still produces competitive qualitative results compared to previous methods. In some cases, especially those with smaller movements, we manage to produce more detailed images as presented in Fig.~\ref{fig:multi-frame}.

\begin{table}[t]
\centering
\small
\vspace{-5pt}
    \begin{tabular}{@{}lcccccc@{}}
        \toprule & \multicolumn{2}{c}{Loss} & \multicolumn{2}{c}{Metrics} \\
        \cmidrule(lr){2-3} \cmidrule(lr){4-5}
         & \(\mathcal{L}_{vol}\) &  \(\mathcal{L}_{tv}\)  & PSNR & MS-SSIM \\
        \midrule
         & - & - & 22.73 & 0.67 \\
        &$\checkmark$ & - & 25.58 & 0.76 \\
         &- & $\checkmark$ & 24.89 & 0.78 \\
         &$\checkmark$ & $\checkmark$ & \textbf{29.51} & \textbf{0.92} \\
        \bottomrule
    \end{tabular}
    \vspace{-5pt}
    \caption{Comparison of PSNR, SSIM for ablation studies with Volume regularization and Total Variation Loss on HyperNeRF-Chicken without gradient-based detection.}
    \label{table:ablations_1}
    \vspace{-10pt}
\end{table}
\begin{table}[h!t]
    \centering
    \small
    
        \begin{tabular}{@{}lcccccc@{}}
        \toprule
         & \multicolumn{2}{c}{Metrics} \\
        \cmidrule(lr){2-3} \cmidrule(lr){4-5}& PSNR & MS-SSIM \\
        \midrule
        Ours w/o. HexPlane & 22.31 & 0.71\\
        Ours w/o. Refinement  & 28.61 & 0.88 \\
        Ours w/o. View dependent sorting & 29.31 & 0.91 \\
        Ours w/o. Gradient-Based Detection & 29.51 & 0.92\\
        Ours & \textbf{31.28 }& \textbf{0.94}\\
        \bottomrule
    \end{tabular}
    \vspace{-5pt}
    \caption{Comparison of PSNR and SSIM for ablations on HyperNeRF-Chicken.}
    \label{table:ablations_2}
    \vspace{-10pt}
\end{table}
\subsection{Ablation Study}
\label{sec:ablation}
\begin{figure}[!t]
    \centering
    \includegraphics[width=1\linewidth]{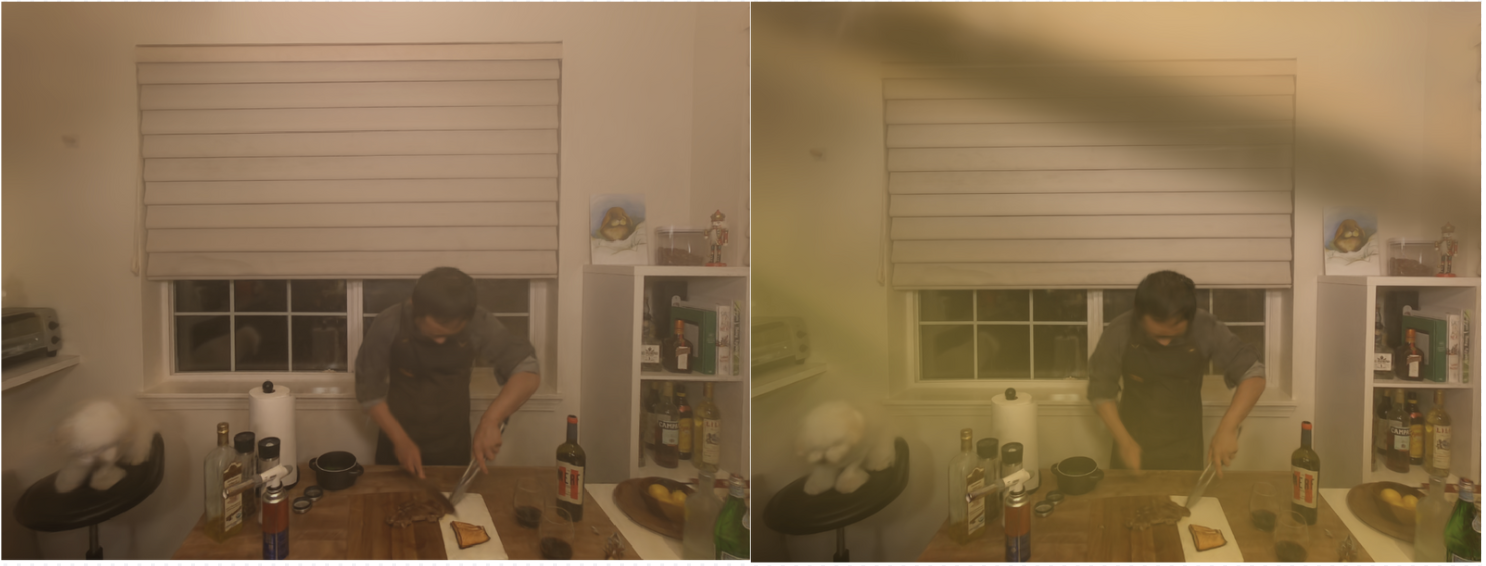}
    \caption{Left: with $L_{vol}$ Right: w/o. $L_{vol}$.}
     \vspace{-20pt}
    \label{fig:scale_compare}
\end{figure}

\noindent\textbf{Loss Design.}
We present the ablation studies in Table \ref{table:ablations_1} to highlight the differences in quantitative results regarding Volume regularization and Total Variation Loss.
We also rendered images specifically focused on Volume regularization, as it had a significant impact on the qualitative results.
As shown in Fig. \ref{fig:scale_compare} without the volume regularization term 
\(\mathcal{L}_{vol}\), we observe large Gaussians floating in front of the camera, causing the blurry effect on the image.

\noindent\textbf{Selective Deformation Strategy.}
A key design choice in our method is the selective deformation of Gaussian properties. Table \ref{table:ablations_3} demonstrates the impact of this strategy. When deforming all properties (including color and opacity), training becomes unstable and quality degrades. Our selective approach, which only deforms geometric properties (position, scale, rotation), maintains stable training while achieving better results.
\begin{table}[t]
    \centering
    \small
    \begin{tabular}{@{}lcccc@{}}
        \toprule
         Deformation Strategy & PSNR & MS-SSIM & Training Stability \\
        \midrule
        All Properties & 26.42 & 0.81 & Unstable \\
        Geometric Only (Ours) & \textbf{29.51} & \textbf{0.92} & Stable \\
        Appearance Only & 24.18 & 0.73 & Stable \\
        No Deformation & 22.31 & 0.71 & Stable \\
        \bottomrule
    \end{tabular}
    \vspace{-5pt}
    \caption{Ablation study on selective deformation strategy. Deforming only geometric properties achieves the best balance between quality and training stability.}
    \label{table:ablations_3}
\end{table} 

\begin{figure}[t!]
    \centering
    \vspace{-10pt}
    \includegraphics[width=0.5\linewidth]{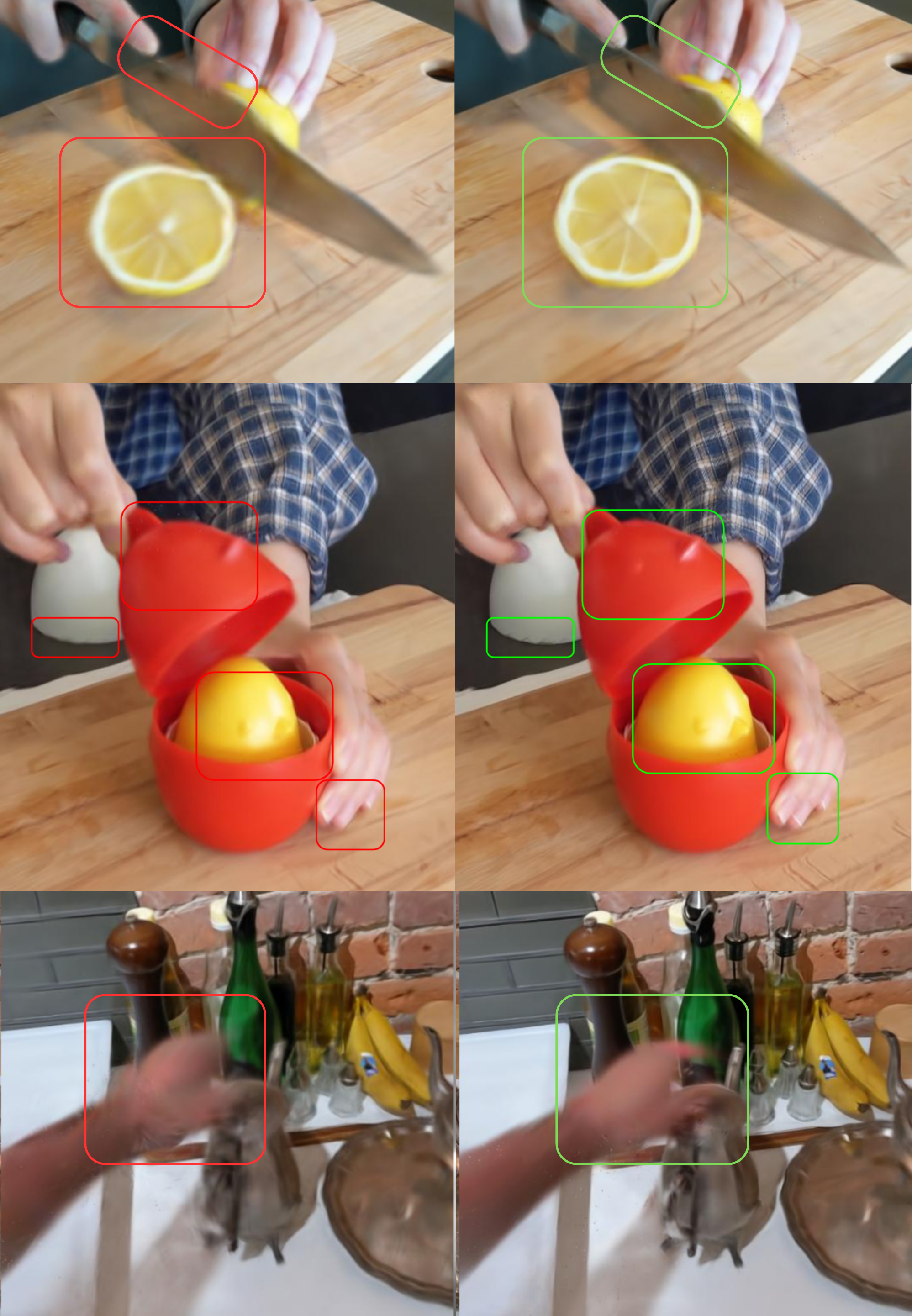}
    \vspace{-5pt}
    \caption{Left: w/o. view refinement. Right:with view refinement. Best viewed at 4$\times$ zoom.}
    \label{fig:adaptive}
    \vspace{-15pt}
\end{figure}

\noindent\textbf{View Refinement.}
We observed that some viewpoints were not reconstructed accurately, which we attribute to inconsistencies in the Gaussians. From Table \ref{table:ablations_2}, the view refinement effectively enhances image quality by refining consistency across viewpoints, as illustrated in Fig. \ref{fig:adaptive}, resulting in more visually appealing outcomes.

\subsection{Memory and Speed Analysis}
\noindent\textbf{Memory Usage.} Our method requires only 3,050 MiB, achieving 86\% reduction vs. TiNeuVox-B (21,558 MiB), 32\% vs. 4D-GS (4,500 MiB), and 56\% vs. GaGS (6,875 MiB). This enables deployment on consumer hardware with limited GPU resources.

\noindent\textbf{Training Time.} Our model completes training in 17 minutes with refinement (13 minutes without), achieving 94\% speedup vs. HyperNeRF (32 hours), 32\% vs. 4D-GS (25 minutes), and 89\% vs. GaGS (120 minutes). This enables rapid deployment in real-time interactive environments.

\vspace{-5pt}
\subsection{Limitations}
Though our approach 
achieves rapid convergence and yields real-time rendering in many scenarios, a few key challenges remain.

\noindent\textbf{Large Motions.} The absence of initial points and inaccuracies in camera poses make it challenging to optimize the Gaussians effectively. Although incorporating view refinement has improved this issue, further enhancements in motion estimation are still required. 

\noindent\textbf{Popping Artifacts.} Although 
the view-dependent sorting solves the ``static'' part of the popping artifact, it isn't entirely eliminated. Further enhancements in voxel representation and 3D Gaussians' rasterizing process are needed to completely remove the artifacts.
\section{Conclusion}
\label{sec:conclusion}
We introduced 4D Neural Voxel Splatting (4D-NVS), a method for real-time dynamic scene rendering that extends voxel representations to the temporal domain. By decoupling spatial structure from temporal dynamics through neural Gaussians and deformation fields, our approach achieves superior rendering quality with memory reduction and computational efficiency. 4D-NVS opens new avenues for real-time dynamic visualization and interactive scene manipulation across various applications.

\noindent\textbf{Compliance with Ethical Standards.} This research was conducted
with the open datasets. It doesn't require ethical approval.

\bibliographystyle{IEEEbib}
\bibliography{strings,ref_update}

\end{document}